\documentclass[10pt,twocolumn,letterpaper]{article}

\usepackage[review]{cvpr}      

\usepackage{graphicx}
\usepackage{amsmath}
\usepackage{amssymb}
\usepackage{booktabs}

\usepackage[pagebackref,breaklinks,colorlinks]{hyperref}

\usepackage[capitalize]{cleveref}
\crefname{section}{Sec.}{Secs.}
\Crefname{section}{Section}{Sections}
\Crefname{table}{Table}{Tables}
\crefname{table}{Tab.}{Tabs.}

\begin{document}

\title{Abstract Art Interpretation Using ControlNet}

\author{Rishabh Srivastava\\
Columbia University\\
New York\\
{\tt\small rs4489@columbia.edu}
\and
Addrish Roy\\
Columbia University\\
New York\\
{\tt\small ar4613@columbia.edu}
}
\maketitle

\begin{abstract}
   Our study delves into the fusion of abstract art interpretation and text-to-image synthesis, addressing the challenge of achieving precise spatial control over image composition solely through textual prompts. Leveraging the capabilities of ControlNet, we empower users with finer control over the synthesis process, enabling enhanced manipulation of synthesized imagery. Inspired by the minimalist forms found in abstract artworks, we introduce a novel condition crafted from geometric primitives such as triangles.
\end{abstract}

\section{Introduction}
\label{sec:intro}

Visual inspiration often strikes unexpectedly, prompting a desire to immortalize fleeting mental imagery into tangible forms. With the advent of text-to-image diffusion models \cite{diffusionModels}, such aspirations have become increasingly attainable through the simple act of textual description. However, the inherent limitations of these models in providing precise spatial control over image composition persist, presenting challenges in accurately conveying complex layouts, poses, shapes, and forms solely through textual prompts.

While existing methodologies primarily rely on textual prompts to guide image generation, the integration of additional image conditions has emerged as a promising approach to empower users with finer control over the synthesis process. One major breakthrough in this domain was the ControlNet architecture \cite{ControlNet}. ControlNet revolutionized the field by enabling the integration of diverse conditioning inputs with Stable Diffusion, ranging from edge maps and human pose skeletons to segmentation maps and depth information. By treating these additional images as conditions for the image generation process, ControlNet paved the way for enhanced spatial control and nuanced manipulation of synthesized imagery.

Building upon the foundation laid by the ControlNet architecture introduced in the previous year, our project focuses on training ControlNet with a new condition. Inspired by the simplicity and versatility of geometric shapes in abstract artworks, our new condition is crafted from geometric primitives like triangles. These primitives are meticulously arranged to approximate the original image in the training dataset, yet imbue it with the essential qualities of abstract artistry. 

The concept stems from the recognition that abstract art often relies solely on shapes to convey its message, emphasizing the power of minimalist forms in eliciting diverse interpretations. Such artistry, characterized by its reliance on geometric elements, serves as the foundation for our new condition image. Moreover, the subjective nature of human imagination further fuels this endeavor, as diverse textual prompts can yield a myriad of representations from the same abstract artwork. Thus, our pursuit is grounded in the rich tapestry of abstract expressionism, where geometric shapes serve as the catalyst for exploring the boundless realm of artistic interpretation and creativity.

In summary, our project presents a pioneering exploration into the intersection of abstract art interpretation and text-to-image synthesis. We aim to train a novel condition and conduct qualitative assessments of the outcomes, condition fidelity, and the overall quality of the generated images.

\section{Related Work}
\label{sec:formatting}

Diffusion models belong to a category of probabilistic generative models. They operate by gradually altering data through noise injection and subsequently learn the ability to reverse this process for generating samples. Sohl-Dickstein and colleagues pioneered Image Diffusion Models \cite{pmlr-v37-sohl-dickstein15}, a concept subsequently adapted to image generation\cite{dm_img_synth}. Latent Diffusion Models (LDM) \cite{diffusionModels} streamline computation by conducting diffusion steps in the latent image space \cite{dm_fast}. Text-to-image diffusion models, utilizing pre-trained language models such as CLIP \cite{dmclip}, excel in generating images by translating textual inputs into latent vectors. Stable Diffusion \cite{stablediff}, a large-scale implementation of latent diffusion, is notable in the field.

Controlling Image Diffusion Models empowers users with the ability to personalize, customize, and generate images tailored to specific tasks. Methods for guiding image generation through text focus on adjusting prompts, manipulating CLIP features, and modifying cross-attention mechanisms \cite{t2iclip}. Approaches like MakeAScene \cite{makescene} encode segmentation masks into tokens to control images, while GLIGEN \cite{gligen} adapts attention layers of diffusion models to learn new parameters for grounded generation. Personalization techniques such as Textual Inversion \cite{textinv} and DreamBooth \cite{db} fine-tune image diffusion models using user-provided example images to customize content in generated images. Tools for prompt-based image editing \cite{predit} offer practical ways to manipulate images with textual prompts.

In image-to-image translation, an image from one domain is transformed into an image from another domain. Conditional Generative Adversarial Networks (GANs) \cite{cgan} and transformers \cite{tf} have demonstrated their capacity to learn mappings between disparate image domains. Manipulating pre-trained GANs allows for addressing specific image-to-image tasks; for example, StyleGANs can be controlled using additional encoders \cite{stygan}.

Unlike image diffusion models that only rely on text prompts for image generation, ControlNet \cite{ControlNet} extends the capabilities of pre-trained large diffusion models to incorporate additional semantic maps, such as edge maps, segmentation maps, key points, shape normals, and depth cues. We delve further into the ControlNet architecture in Section 3.2.

\section{Methodology}
\subsection{Dataset Preparation}

\begin{figure}[t]
  \centering
  \includegraphics[width=\linewidth]{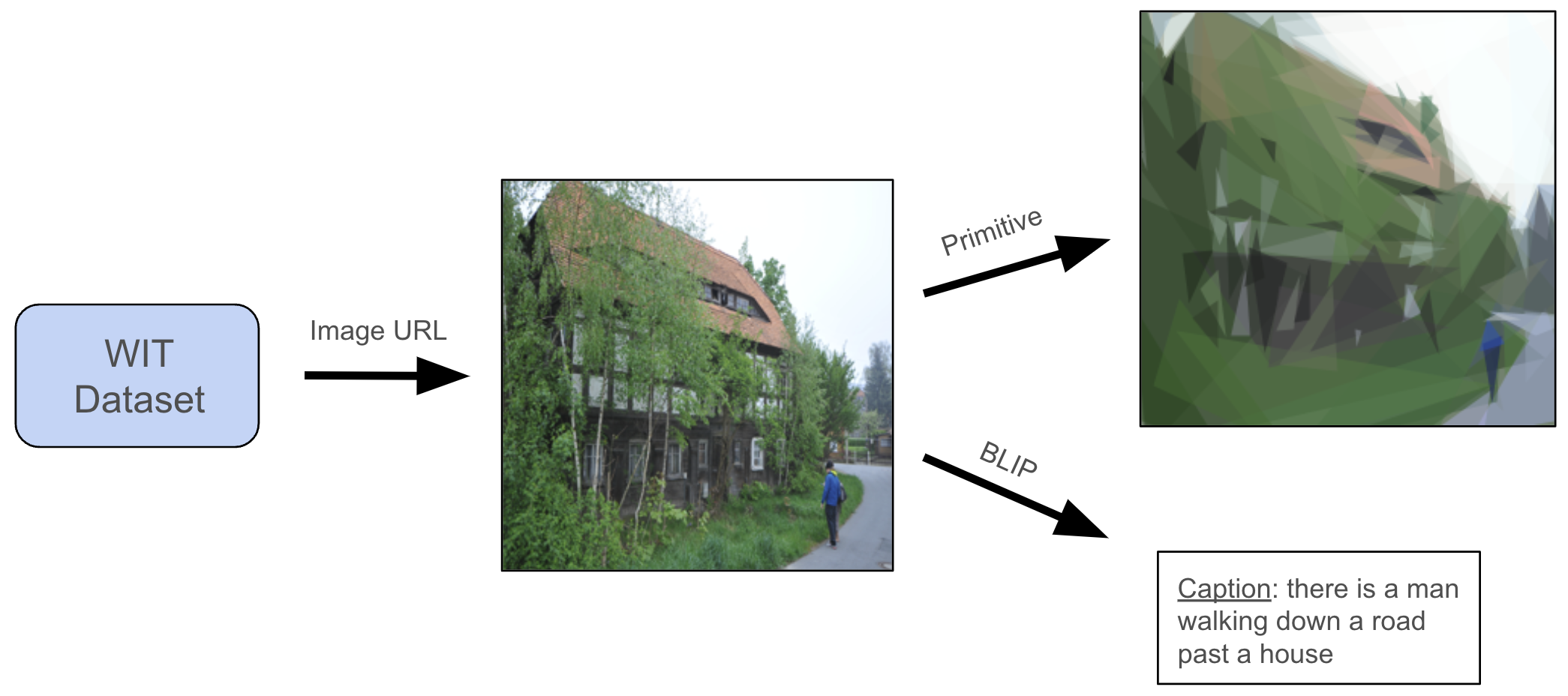}

   \caption{Representation of our dataset generation pipeline.}
   \label{fig:dataset}
\end{figure}

The dataset utilized in this study was meticulously curated from a subset of the Wikipedia-based Image Text (WIT) Dataset \cite{wit}, a colossal multimodal multilingual repository comprising an astounding 37.6 million image-text examples.

To construct our dataset, we utilized the image URLs sourced from the WIT dataset to download the original images. These images served as the target images when we trained our ControlNet model. While the WIT dataset offers image descriptions, their specificity and granularity were not ideally suited for training our model. Thus, captions for the images were generated using Hugging Face's implementation of the Bootstrapping Language-Image Pre-training for Unified Vision-Language Understanding and Generation (BLIP) model \cite{blip}.

In tandem with obtaining images and captions, control images were derived from the downloaded images using the innovative Primitive software \cite{primitive}. Employing an iterative methodology, Primitive progressively refines a target image by adding geometric shapes to minimize the disparity between the target and the drawn image. This iterative refinement process continues until a visually appealing yet abstract representation is achieved, imbuing the dataset with a nuanced understanding of artistic abstraction and creativity. The current dataset uses 50 triangles to approximate the target image.

In summary, our dataset \cite{myds} comprises 14,279 pairs of control and target images, each accompanied by meticulously crafted captions. By leveraging the inherent variability and complexity within the dataset, ControlNet can be trained to adeptly manipulate stable diffusion images based on newly introduced conditions.

\subsection{ControlNet Architecture}
\begin{figure}[t]
  \centering
  \includegraphics[width=\linewidth]{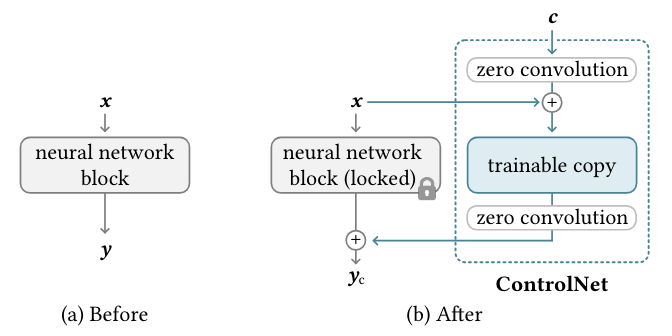}

   \caption{To incorporate a ControlNet into the block illustrated in (a), the original block is locked and a trainable duplicate is generated as the first step. These two blocks are then linked using zero convolution layers, specifically employing 1 $\times$ 1 convolutions with weight and bias parameters set to 0. Here, the conditioning vector $c$ represents additional information we aim to integrate into the network, as depicted in (b).}
   \label{fig:controlnet}
\end{figure}


Fig. \ref{fig:controlnet} above helps to visualize the setup for ControlNet where we can see that ControlNet integrates additional conditions into the main neural network block (for example, it can be resnet block, multi-head attention block, transformer block, etc.). Consider the trained neural network block to be represented by $\mathit{F(\cdot; \Theta)}$ where the $\Theta$ refers to the parameters of the block. Let, $\mathbf{x}$ \& $\mathbf{y}$ denote the input \& output of this block. Then, 
\begin{equation}
    \mathbf{y}=\mathit{F(\mathbf{x}; \Theta)}
\end{equation}
To inject a ControlNet into this block, we lock the parameters $\Theta$ of the pre-trained original block and duplicate the block to create a trainable copy with parameters $\Theta_c$. The input to the trainable copy is a conditioning vector, $\mathbf{c}$ which comes from the outside. Let $\mathit{Z(\cdot; \cdot)}$ denote the zero convolution layers that connect the trainable copy to the locked model. Two instances of such zero convolutions are used with parameters, $\Theta_{z1}$ \& $\Theta_{z2}$ respectively. Then, the output of the ControlNet can be represented by
\begin{equation}
    y_c=F(x; \Theta) + Z\left(F\left(x + Z(c; \Theta_{z1}); \Theta_{c}\right); \Theta_{z2}\right)
\end{equation}
After the 1st training step, the above equation (2) reduces to the equation below:
\begin{equation}
    \mathbf{y_c}=\mathbf{y}
\end{equation}
Thus, when the training starts harmful noise cannot impact the hidden states of the network block in the trainable copy. Additionally, given that $Z(c; \Theta_{z1}) = 0$ and the trainable copy also takes the input image $x$, it remains fully operational and preserves the capabilities of the extensive, pre-trained model. This enables it to function as a robust foundation for continued learning. 

\subsection{Training}


Once the dataset is prepared, the next step is to load the pre-trained model. The model architecture is defined in a YAML configuration file, encapsulating its structure and parameters. Loading the pre-trained model weights initializes the model with prior knowledge learned from extensive training on large datasets. This initialization kickstarts the model's ability to generate meaningful images based on textual input.

With the dataset and model in place, the training process is initialized. This involves setting up a DataLoader to efficiently feed batches of data into the model during training. The DataLoader handles tasks such as batching, shuffling, and parallel data loading, optimizing the training workflow.

The model is then trained to learn to generate images by iteratively processing image-text pairs and optimizing its parameters to minimize the discrepancy between generated and ground truth images like in Stable Diffusion. During each iteration, the model adjusts its internal representations, gradually improving its ability to translate textual prompts into realistic images. The use of GPU acceleration speeds up this process significantly, allowing for efficient exploration of the high-dimensional parameter space.

\subsection{Inference}

Once the ControlNet parameters are trained, the model becomes capable of generating new images. Inference becomes straightforward by inputting a sample abstract image along with a prompt. While inference can technically function without prompts, it is important to note that our model's performance without prompts may be sub-optimal. This limitation arises from the fact that our model has not been extensively trained on datasets lacking prompts.

\section{Experimental Results}

\begin{figure*}[t]
\centering
\includegraphics[width=\linewidth]{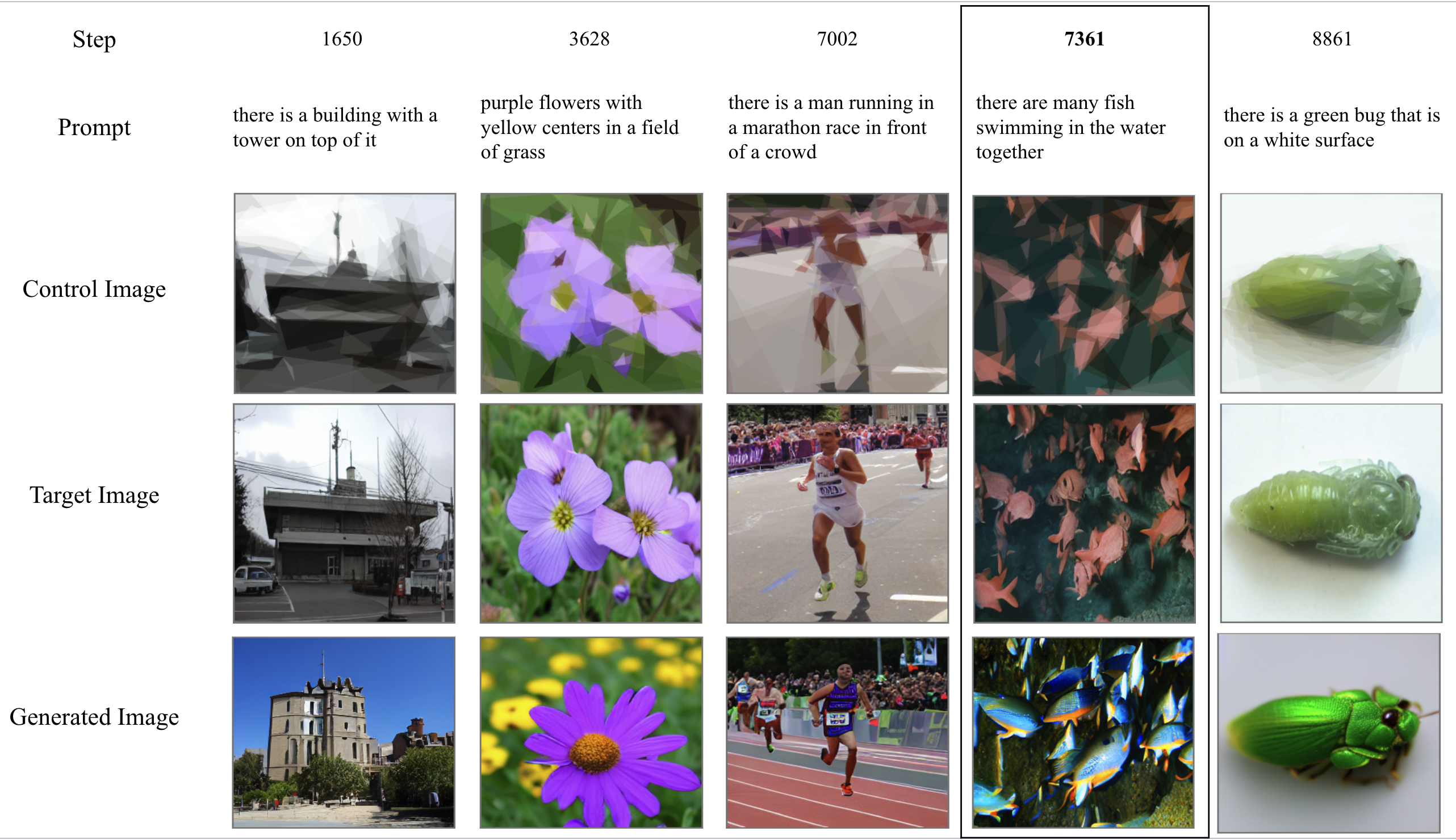}
\caption{Sudden convergence phenomenon: ControlNet consistently produces high-quality images throughout training, with a marked instance (e.g., bolded step 7361) where it abruptly starts intersecting with the control image. Although the intersection of the generated image with the target image increases at step 7361, the colors do not match completely.}
\label{fig:sconv}
\end{figure*}

\begin{figure*}[t]
\centering
\includegraphics[width=\linewidth]{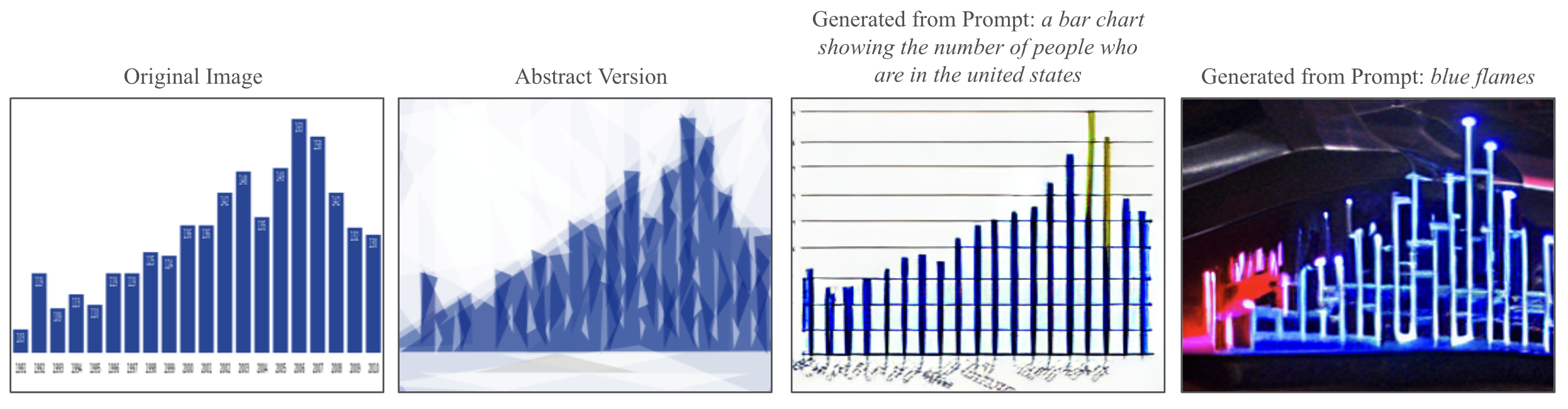}
\caption{An example of how the same abstract image can be interpreted differently.}
\label{fig:sabs}
\end{figure*}

We applied the ControlNet architecture on Stable Diffusion and trained the resultant model using 14000 images from our dataset. The model was trained on an Nvidia T4 GPU, and a batch size of 2 was employed to ensure compliance with the memory constraints of the GCP Compute Engine instance. The code can be found at \cite{coderep}.

As stated in the ControlNet paper \cite{ControlNet}, we also observed the sudden convergence phenomenon (Fig. \ref{fig:sconv}). As a result of zero convolutions, ControlNet consistently generates high-quality images throughout the training process. However, at a specific stage in training, the model undergoes a sudden learning event wherein it begins to adhere closely to the input condition, which is noted as the sudden convergence phenomenon.

Fig. \ref{fig:sabs} illustrates the results generated by our trained model for a given test image. This same image can be perceived in multiple ways, as evidenced by the diverse prompts showcased. Each prompt breathes life into its own distinct imaginative space, showcasing the richness and depth of creative interpretation.

\section{Discussion}

Successfully training a ControlNet model on a novel control image concept marks a significant achievement in our research endeavor. To facilitate this, we meticulously curated a custom dataset tailored to our specific requirements.

During the training process, our ControlNet exhibited remarkable adaptability and proficiency, effectively learning to generate diverse images in accordance with the provided prompts. Notably, it adeptly preserved the approximated object locations, underscoring its ability to maintain spatial consistency while exploring the creative possibilities inherent in the input conditions.

However, a notable limitation was observed: while the trained model exhibited proficiency in determining object placements and aligning with the desired prompt descriptions and input control image, it struggled to accurately replicate colors sometimes. This shortfall likely stems from the need for further training. Regrettably, due to constraints on compute resources, we were unable to conduct more extensive training sessions to address this deficiency.

\section{Conclusion}

In conclusion, our exploration delves into the confluence of abstract art interpretation and text-to-image synthesis, addressing the intrinsic challenges of achieving precise spatial control over image composition solely through textual prompts. By extending the capabilities of pre-trained large diffusion models, ControlNet empowers users with finer control over the synthesis process, laying the groundwork for enhanced spatial manipulation of synthesized imagery.

Drawing inspiration from the minimalist forms found in abstract artworks, our project introduces a novel condition crafted from geometric primitives, such as triangles. Rooted in the recognition of abstract art's reliance on shapes to convey its message, our endeavor harnesses the subjective nature of human imagination, where diverse textual prompts yield myriad representations from the same abstract artwork.

Despite notable proficiency in preserving object locations and aligning with desired prompts, our trained ControlNet model exhibited limitations in accurately replicating colors, likely necessitating further training to address this deficiency, hampered by compute resource constraints.

In our future work, we aim to enhance our dataset by incorporating a wider variety of geometric shapes and increasing the number of shapes used to approximate different images. We also aim to explore evaluating the synthesized images quantitatively. 

{\small
\bibliographystyle{ieee_fullname}
\bibliography{egbib}

\begin{thebibliography}{10}\itemsep=-1pt

\bibitem{predit}
Tim Brooks, Aleksander Holynski, and Alexei~A. Efros.
\newblock Instructpix2pix: Learning to follow image editing instructions.
\newblock \textit{arXiv preprint arXiv:2211.09800}, 2023.

\bibitem{cgan}
Y. Choi, M. Choi, M. Kim, J. Ha, S. Kim, and J. Choo.
\newblock Stargan: Unified generative adversarial networks for multi-domain image-to-image translation.
\newblock In {\em 2018 IEEE/CVF Conference on Computer Vision and Pattern Recognition (CVPR)}, pages 8789--8797, jun 2018.

\bibitem{dm_img_synth}
Prafulla Dhariwal and Alexander Nichol.
\newblock Diffusion models beat gans on image synthesis.
\newblock In {\em Advances in Neural Information Processing Systems}, volume~34, pages 8780--8794. Curran Associates, Inc., 2021.

\bibitem{dm_fast}
Patrick Esser, Robin Rombach, and Bjorn Ommer.
\newblock Taming transformers for high-resolution image synthesis.
\newblock In {\em Proceedings of the IEEE/CVF Conference on Computer Vision and Pattern Recognition (CVPR)}, pages 12873--12883, June 2021.

\bibitem{tf}
P. Esser, R. Rombach, and B. Ommer.
\newblock Taming transformers for high-resolution image synthesis.
\newblock In {\em 2021 IEEE/CVF Conference on Computer Vision and Pattern Recognition (CVPR)}, pages 12868--12878, jun 2021.

\bibitem{primitive}
Michael Fogleman.
\newblock Primitive.
\newblock \textit{https://github.com/fogleman/primitive}, 2016.

\bibitem{makescene}
Oran Gafni, Adam Polyak, Oron Ashual, Shelly Sheynin, Devi Parikh, and Yaniv Taigman.
\newblock Make-a-scene: Scene-based text-to-image generation with human priors.
\newblock In {\em Computer Vision – ECCV 2022: 17th European Conference, Tel Aviv, Israel, October 23–27, 2022, Proceedings, Part XV}, page 89–106, 2022.

\bibitem{textinv}
Rinon Gal, Yuval Alaluf, Yuval Atzmon, Or Patashnik, Amit~H. Bermano, Gal Chechik, and Daniel Cohen-Or.
\newblock An image is worth one word: Personalizing text-to-image generation using textual inversion.
\newblock \textit{arXiv preprint arXiv:2208.01618}, 2022.

\bibitem{blip}
Junnan Li, Dongxu Li, Caiming Xiong, and Steven C.~H. Hoi.
\newblock Blip: Bootstrapping language-image pre-training for unified vision-language understanding and generation.
\newblock In {\em International Conference on Machine Learning}, 2022.

\bibitem{gligen}
Yuheng Li, Haotian Liu, Qingyang Wu, Fangzhou Mu, Jianwei Yang, Jianfeng Gao, Chunyuan Li, and Yong~Jae Lee.
\newblock Gligen: Open-set grounded text-to-image generation.
\newblock \textit{arXiv preprint arXiv:2301.07093}, 2023.

\bibitem{dmclip}
Alec Radford, Jong~Wook Kim, Chris Hallacy, Aditya Ramesh, Gabriel Goh, Sandhini Agarwal, Girish Sastry, Amanda Askell, Pamela Mishkin, Jack Clark, Gretchen Krueger, and Ilya Sutskever.
\newblock Learning transferable visual models from natural language supervision.
\newblock In {\em Proceedings of the 38th International Conference on Machine Learning}, volume 139 of {\em Proceedings of Machine Learning Research}, pages 8748--8763. PMLR, 18--24 Jul 2021.

\bibitem{t2iclip}
Aditya Ramesh, Prafulla Dhariwal, Alex Nichol, Casey Chu, and Mark Chen.
\newblock Hierarchical text-conditional image generation with clip latents.
\newblock \textit{arXiv preprint arXiv:2204.06125}, 2022.

\bibitem{stygan}
Elad Richardson, Yuval Alaluf, Or Patashnik, Yotam Nitzan, Yaniv Azar, Stav Shapiro, and Daniel Cohen-Or.
\newblock Encoding in style: a stylegan encoder for image-to-image translation.
\newblock In {\em 2021 IEEE/CVF Conference on Computer Vision and Pattern Recognition (CVPR)}, pages 2287--2296, 2021.

\bibitem{diffusionModels}
R. Rombach, A. Blattmann, D. Lorenz, P. Esser, and B. Ommer.
\newblock High-resolution image synthesis with latent diffusion models.
\newblock In {\em 2022 IEEE/CVF Conference on Computer Vision and Pattern Recognition (CVPR)}, pages 10674--10685, 2022.

\bibitem{db}
Nataniel Ruiz, Yuanzhen Li, Varun Jampani, Yael Pritch, Michael Rubinstein, and Kfir Aberman.
\newblock Dreambooth: Fine tuning text-to-image diffusion models for subject-driven generation.
\newblock \textit{arXiv preprint arXiv:2208.12242}, 2023.

\bibitem{pmlr-v37-sohl-dickstein15}
Jascha Sohl-Dickstein, Eric Weiss, Niru Maheswaranathan, and Surya Ganguli.
\newblock Deep unsupervised learning using nonequilibrium thermodynamics.
\newblock In {\em Proceedings of the 32nd International Conference on Machine Learning}, volume~37 of {\em Proceedings of Machine Learning Research}, pages 2256--2265, Lille, France, 07--09 Jul 2015. PMLR.

\bibitem{wit}
Krishna Srinivasan, Karthik Raman, Jiecao Chen, Michael Bendersky, and Marc Najork.
\newblock Wit: Wikipedia-based image text dataset for multimodal multilingual machine learning.
\newblock In {\em Proceedings of the 44th International ACM SIGIR Conference on Research and Development in Information Retrieval}, SIGIR ’21. ACM, July 2021.

\bibitem{myds}
Rishabh Srivastava.
\newblock geometricshapes14k.
\newblock \textit{https://www.kaggle.com/datasets/rishabhsrivastava66/images-made-up-of-geometric-shapes-controlnet/data}, 2024.

\bibitem{coderep}
Rishabh Srivastava and Addrish Roy.
\newblock Abstract-art-interpretation-using-controlnet.
\newblock \textit{https://github.com/RishabhS66/Abstract-Art-Interpretation-Using-ControlNet}, 2024.

\bibitem{stablediff}
Stability.
\newblock Stable diffusion v1.5 model card, https://huggingface.co/runwayml/stable-diffusion-v1-5, 2022.

\bibitem{ControlNet}
Lvmin Zhang, Anyi Rao, and Maneesh Agrawala.
\newblock Adding conditional control to text-to-image diffusion models.
\newblock In {\em 2023 IEEE/CVF International Conference on Computer Vision (ICCV)}, pages 3813--3824, 2023.

\end{thebibliography}
}

\end{document}